\documentclass[twoside,leqno,twocolumn]{article}

\usepackage[letterpaper]{geometry}

\usepackage{ltexpprt}
\usepackage{hyperref}
\usepackage{graphicx}
\usepackage{float}
\usepackage{xcolor}

\begin{document}

\newcommand\relatedversion{}

\title{\Large A Scalable Data-Driven Framework for Systematic Analysis of SEC 10-K Filings Using Large Language Models\relatedversion}
\author{Syed Affan Daimi\thanks{International Institute of Information Technology, Hyderabad}
\and Asma Iqbal\thanks{Deccan College of Engineering And Technology}}

\date{}

\maketitle




\fancyfoot[R]{\scriptsize{Copyright \textcopyright\ 2024\\
Copyright for this paper is retained by authors}}



\begin{abstract} \small\baselineskip=9pt The number of companies listed on the NYSE has been growing exponentially, creating a significant challenge for market analysts, traders, and stockholders who must monitor and assess the performance and strategic shifts of a large number of companies regularly. There is an increasing need for a fast, cost-effective, and comprehensive method to evaluate the performance and detect and compare many companies' strategy changes efficiently. We propose a novel data-driven approach that leverages large language models (LLMs) to systematically analyze and rate the performance of companies based on their SEC 10-K filings. These filings, which provide detailed annual reports on a company’s financial performance and strategic direction, serve as a rich source of data for evaluating various aspects of corporate health, including confidence, environmental sustainability, innovation, and workforce management. We also introduce an automated system for extracting and preprocessing 10-K filings. This system accurately identifies and segments the required sections as outlined by the SEC, while also isolating key textual content that contains critical information about the company. This curated data is then fed into Cohere's Command-R+ LLM to generate quantitative ratings across various performance metrics. These ratings are subsequently processed and visualized to provide actionable insights. The proposed scheme is then implemented on an interactive GUI as a no-code solution for running the data pipeline and creating the visualizations. The application showcases the rating results and provides year-on-year comparisons of company performance.\end{abstract}

\section{Introduction.}In the fast-paced world of modern business, stakeholders such as market analysts, investors, and decision-makers face the challenge of effectively assessing and comparing the performance of numerous companies. Traditionally, this process involves sifting through extensive financial filings and qualitative reports, often requiring substantial time and expertise. While these traditional methods provide valuable insights, they typically result in narrative analyses rather than easily digestible metrics, making it difficult to quickly gauge and compare the performance of multiple companies.

Moreover, the sheer volume of companies listed on major stock exchanges, such as the NYSE, exacerbates the problem. Conducting detailed analyses on each company is not only time-consuming but also expensive, limiting the feasibility of such assessments on a large scale. This has created a need for a more efficient, scalable “litmus test” solution that can deliver quick, quantifiable insights into company performance.

In recent years, the finance industry has increasingly turned to artificial intelligence (AI) and large language models (LLMs) to enhance decision-making processes and provide sophisticated investment advice. Robo-advisors have emerged as one of the most prominent applications of AI in finance, revolutionizing wealth management and investment advisory services. These platforms leverage LLMs to analyze vast amounts of financial data, enabling them to offer personalized investment strategies and continuously adjust portfolios based on market dynamics and individual risk preferences \cite{Ahmed2022}. The ability of LLMs to parse complex data sets and identify intricate market patterns has allowed robo-advisors to provide more informed and adaptive investment guidance compared to traditional methods.

The integration of LLMs into robo-advisory platforms has not only improved the accuracy and relevance of investment advice but has also made such services more accessible to a broader audience. For instance, AI-driven models have been employed to optimize portfolio construction and ESG investment targets, as well as to recommend peer-to-peer (P2P) loan investments \cite{Ashta2021}. 

However, these systems have yet to address the broader need for scalable, generalized methods that can systematically evaluate and compare the performance of a large number of companies across diverse sectors. Building on the success of AI in investment advisory services, there is growing interest in applying LLMs to other areas of financial decision support, including the analysis of corporate financial documents. Traditional financial analysis methods often result in narrative-heavy insights, which can be time-consuming to interpret and difficult to compare across multiple companies. 

Research has shown that LLMs can perform effectively in data-limited scenarios, such as banking intent detection, using few-shot learning techniques \cite{Loukas2023}. By leveraging the analytical power of LLMs, our novel approach aims to address these limitations by introducing a system that automates the extraction and processing of 10-K filings, generating numeric ratings that provide clear, comparative measures of company performance. These ratings provide a clear, comparative measure of a company's performance, allowing stakeholders to quickly assess and compare companies across a broad range of criteria. By leveraging large language models (LLMs) such as Cohere's command-R+ to generate these ratings, our method is both time-efficient and cost-effective, enabling the analysis of a large number of companies with unprecedented speed and accuracy.

Furthermore, conducting a year-on-year comparison of the same company mitigates the biases that arise when comparing entities operating in different contexts. This longitudinal analysis uncovers trends, strengths, and areas for improvement specific to each company, offering a more nuanced understanding of its strategic evolution and operational trajectory. By focusing on intrinsic performance changes rather than external comparisons, our approach ensures a more accurate and fair assessment, highlighting the company’s progress over time. To facilitate this process, we have developed a web application that serves as a no-code solution, enabling the end-user to effortlessly implement our system, analyze data, and visualize results without requiring technical expertise.

\section{Related Works.}
\subsection{Large Language Models (LLMs) for Sentiment Analysis.}

Sentiment analysis, the process of identifying and extracting subjective information from text, has become increasingly important in fields such as social media monitoring, customer service, and political discourse analysis. Traditional approaches to sentiment analysis often relied on rule-based systems or machine learning models trained on limited datasets \cite{Rajapaksha2020}. However, the emergence of LLMs has brought about a paradigm shift, allowing for more accurate and scalable sentiment analysis \cite{Cheng2024} \cite{Chun2023}. 

LLMs excel at sentiment analysis due to their ability to capture complex linguistic patterns, understand context, and reason about emotions \cite{Tornberg2023}. By leveraging their broad knowledge and in-context learning capabilities, LLMs can effectively identify and classify sentiment in various types of text, such as social media posts, product reviews, and news articles \cite{Alderazi2024}. 

Recent research has focused on enhancing LLMs' performance in sentiment analysis through the use of prompting strategies \cite{Wang2023}. These strategies involve carefully crafting input prompts that guide the model to generate more accurate and relevant sentiment predictions. Prompting techniques such as role-playing and chain-of-thought have shown promising results in improving LLMs' ability to handle complex sentiment analysis tasks, particularly in identifying implicit emotions and sentiment in ambiguous contexts. 

Moreover, LLMs have been explored for targeted sentiment analysis, which focuses on identifying sentiment towards specific entities or aspects within a text \cite{Juros2024}. This is particularly relevant in the analysis of news headlines, where entities are often portrayed in specific ways to evoke sentiment \cite{Alderazi2024}. LLMs have demonstrated the ability to effectively handle targeted sentiment analysis, outperforming fine-tuned encoder models in various languages.

As LLMs continue to advance, their applications in sentiment analysis are expected to grow, enabling more accurate and efficient analysis of large-scale textual data.
 
\subsection{LLMs for Textual Analysis.} 

Large Language Models (LLMs) have emerged as powerful tools for textual analysis, significantly enhancing various natural language processing (NLP) tasks, including document question answering, retrieval-augmented generation (RAG), and text summarization. Their capabilities are particularly beneficial in specialized domains such as finance.

\textbf{Document Question Answering:} LLMs excel in document question answering by leveraging vast amounts of text data to provide accurate and contextually relevant answers \cite{Zhu2024}. For instance, the development of frameworks like UniGen integrates generative document retrieval and grounded answer generation into a unified model. This approach allows for simultaneous optimization of retrieval and question-answering tasks, demonstrating superior performance on datasets like MS MARCO and NQ \cite{Li2024}. In finance, systems like SEC-QA have been proposed to automate the analysis of long financial documents, generating question-answer pairs that reflect real-world scenarios and improving the accuracy of responses through continuous dataset updates \cite{Lai2024}.

\textbf{Retrieval-Augmented Generation (RAG):} RAG enhances the performance of LLMs by retrieving relevant information from external databases during the generation process. This method addresses the limitations of LLMs, which may not have access to the most current data. \cite{Hui2024} Recent developments focus on optimizing the retrieval process and improving performance across various tasks, including text summarization and dialogue generation \cite{Wang2024}. RAG systems have been developed to tackle complex multi-document questions, significantly improving the accuracy of financial data analysis and reasoning tasks \cite{Zhang2024}.

\textbf{Text Summarization:} LLMs are also adept at text summarization, condensing lengthy documents into concise summaries while preserving essential information. Recent advancements have shown that RAG can be particularly effective in this area. For instance, an LLM-based approach demonstrated an 11\% improvement in summarization tasks compared to baseline methods \cite{Wang2024}. Tools like FloodBrain generate comprehensive reports on flood disaster impacts by synthesizing information from various sources, illustrating how LLMs can streamline reporting processes in finance and beyond \cite{Colverd2023}.

\section{Our Methodology.}
Figure 1 is a graphical representation of our end-to-end pipeline which can be classified into three verticals - 1) Data Collection, 2) Data Cleaning and 3) LLM Evaluation. The subsequent sections explain in detail each of the three verticals. Section 3.1 discusses Data Collection and Section 4 delves into Data Cleaning. Lastly, Section 5 discusses LLM Evaluation.

\begin{figure*}[h]
    \centering
    \includegraphics[width=0.8\textwidth]{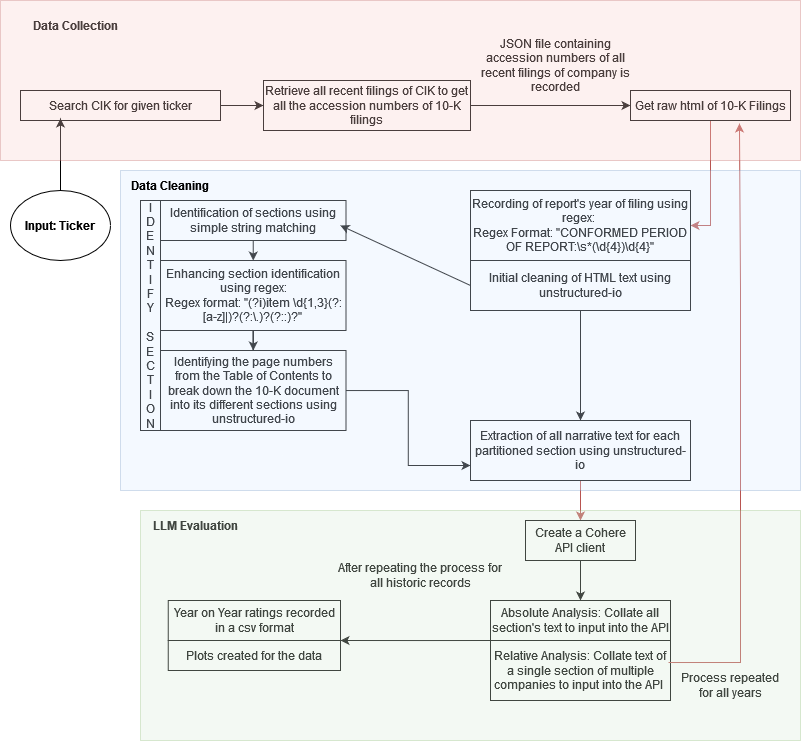}
    \caption{Framework Overview of Proposed System}
\end{figure*}

\subsection{Data Collection.}

To analyze the quality of 10-K filings in our study, we collected data directly from the U.S. Securities and Exchange Commission (SEC) Electronic Data Gathering, Analysis, and Retrieval (EDGAR) database. This process involved querying the SEC-EDGAR endpoints (refer to Table 1 for endpoint details) to obtain comprehensive annual reports filed by public companies, which provide detailed insights into their financial performance and business strategies.

We employed a custom script to systematically retrieve the 10-K filings for a specified list of companies based on their ticker symbols. First, we converted the ticker symbol of each company into its corresponding Central Index Key (CIK), a unique identifier used by the SEC to track corporations and individuals who have filed disclosures. Using the CIK, we queried the SEC-EDGAR database to retrieve a list of all filings associated with that entity. Each filing in the EDGAR system is associated with a unique accession number, which is automatically assigned upon submission. This accession number was crucial in fetching the complete filing text and identifying and filtering for all the historic 10-K filings. 

\begin{table*}[h]
    \centering
    \begin{tabular}{|c|c|}
        \hline
        \textbf{Endpoint} & \textbf{Retrieval} \\ \hline
        sec.gov/cgi-bin/browse-edgar/\{ticker\} & Search CIK for given ticker \\ \hline
        data.sec.gov/submissions/CIK\{CIK\} & Retrieve all recent filings of CIK to get all 10-K filings' accession numbers \\ \hline
        sec.gov/Archives/edgar/data & Get raw HTML of 10-K Filings \\ \hline
    \end{tabular}
    \caption{SEC-EDGAR API Endpoints and their functionalities }
    \label{tab:my_label}
\end{table*}

\section{Data Cleaning.}
The raw HTML data obtained from the 10-K filings required a thorough cleaning process to ensure it was structured for meaningful analysis. Given the unstructured nature of these filings, our data cleaning methodology focused on extracting and organizing the narrative content into the same sections as required by the SEC in their filings, which are crucial for evaluating the qualitative aspects of the filings, and ensuring the limited context window of the LLM is utilized fully. 

\subsection{Date Identification and Demarcation.} 
A critical aspect of the data cleaning process involved extracting the year of each filing to facilitate chronological analysis. This was accomplished by parsing the text content of the filings and employing regular expressions to locate the reporting period. We exploited the uniform formatting guidelines as prescribed by the SEC to create the 10-K filings.

\subsection{Narrative Text Identification.} 
The main objective of the data cleaning process is the identification and extraction of narrative text elements, and tagging each string of the narrative text to the section it belongs to. Narrative text is defined as excerpts which contain important information about the company’s operations. Identification of narrative texts was achieved by using the unstructured-io library, and is a crucial step to ensure maximum utilization of the 128K context window of Cohere’s Command-R family of LLMs. Unstructured-io is a library for data cleaning specifically for downstream LLM tasks. We use unstructured-io to clean the HTML elements and implement Parts of Speech (POS) tagging, to differentiate between narrative text, lists, titles, and other text elements. To qualify as narrative text, the excerpt must contain a significant number of verbs and non-proper nouns.

\subsection{Sections’ Partition Creation.} 
Extraction of labelled text information using predefined structures common in financial documents is a common paradigm \cite{Bentabet2020} \cite{Yepes2024} We also introduce a pipeline designed to identify the various subheadings or sections in the filing, which utilizes predefined document structures and section headings. The pipeline identifies key sections such as "Risk Factors," "Management's Discussion and Analysis," and other relevant parts based on their standard titles and positions within the document, exploiting that the extracted content aligns with the expected format and structure of a typical 10-K filing. Section identification is particularly important to perform relative analysis. (see section 5.2)

Furthermore, we improved our cleaning pipeline by including custom regex-based methods on top of simple string matching to capture any sections or subsections that might not conform strictly to the predefined structures. This flexibility was crucial for handling variations across different companies' filings, as not all filings adhere uniformly to the standard format. 

To partition the narrative text into their corresponding sections, we identify the sections from the “Table of Content” of the document, and extract their page number. Narrative texts belonging to the corresponding page numbers are tagged with their respective sections. 

Once the narrative text was identified and correctly tagged with its corresponding section, the cleaned data was then structured into an Initial Structured Document (ISD) format. This format allowed for the consistent representation of each section, ensuring that the text was organized and labeled according to the specific sections defined by the SEC filing structure.

Finally, the structured data was exported into a CSV format, where each row represents a specific section or subsection, along with its corresponding line of narrative text. Table 2 contains a few example lines of the dataset output at the end of this step.

\begin{table*}[h]
    \centering
    \begin{tabular}{|c|c|p{10cm}|}
        \hline
        \textbf{Section} & \textbf{Element Type} & \textbf{Text} \\ \hline
        BUSINESS & NarrativeText & Payable Metal: Ounces or pounds of metal in concentrate payable to the operator after deducting a percentage of metal in concentrate paid to a third-party smelter pursuant to smelting contracts. \\ \hline
        BUSINESS & NarrativeText & Reserve: That part of a mineral deposit that could be economically and legally extracted or produced at the time of the reserve determination. \\ \hline
        BUSINESS & NarrativeText & Royalty: The right to receive a percentage or other denomination of mineral production from a resource extraction operation. \\ \hline
    \end{tabular}
    \caption{First few lines of the output file for Royal Gold (\$RGLD)}
    \label{tab:output_file_rgold}
\end{table*}

\section{LLM Evaluation.}
After extracting the narrative text, it is fed into an LLM instructed to rate its quality and LLM capabilities to act as text evaluators has been explored \cite{Chen2023} \cite{WangJ2023}. We utilize Cohere's Command-R+ model as our LLM, which is the first open-weight LLM to beat GPT-4 in the Chatbot arena \cite{Chiang2024} \cite{Cohere2023}. Yi Chen et. al, \cite{Chen2023} suggests two scoring techniques for evaluating the quality of a given text using an LLM, pairwise comparison to assess text quality by directly comparing a pair of generated texts, and individual scoring to directly generate a score to measure the absolute quality of each text individually. An important parameter to improve the performance of the scores is the prompt. A prompt is an input to a Generative AI model that is used to guide its output \cite{Schulhoff2024}. We utilize role based zero-shot prompting as inspired by Wu et al \cite{Wu2023}, and Chan et. al [26]. 

\subsection{Absolute Analysis} 
We utilize individual scoring to generate and present yearly ratings of the company in four key dimensions: confidence, environment, innovation, and people. The LLM assumes the role of an AI grader \cite{Wu2023} \cite{Chan2023} to assign a score between 0-2 based on the criterion given (see table 4 for the list of criterions). Additionally, we introduce two “versions” of the AI Grader role, where one AI grader is prompted to be especially strict in its rating in an attempt to reduce bias as attempted by \cite{Chan2023}.

For each dimension, we developed custom grading scripts that interacted with the LLM. For instance, to evaluate the confidence rating, the script passed the narrative text along with a criterion that asked the LLM to assess the company's confidence in its future growth and financial stability. The LLM then returned a score, with 2 indicating strong confidence, 1 representing moderate confidence, and 0 signifying a lack of confidence or financial uncertainty.

Similarly, the environment rating script focused on the company's commitment to sustainability. The narrative was graded based on the presence of actionable environmental plans, commitments to sustainability, or the absence of any mention of environmental responsibility. The LLM's score helped quantify the company's stance on environmental issues as portrayed in its filings.

The innovation rating script assessed the company's focus on technological advancements and future innovations. The LLM evaluated whether the company demonstrated a commitment to innovation through specific actions and plans or merely mentioned it without concrete evidence. The grading criterion distinguished between robust innovation strategies and superficial mentions.

Lastly, the people rating script gauged the importance the company placed on its workforce and talent. The LLM rated the narrative on how well it acknowledged the role of employees and the presence of actionable plans to attract and retain talent, ensuring employee welfare.

Table 3, 4 and 5 contain details of the prompts we utilized for absolute analysis.

\begin{table}[h]
    \centering
    \begin{tabular}{|p{8cm}|}
        \hline
        \textbf{Base Prompt for Absolute Analysis} \\ \hline
        You are an AI grader that, given an output and a criterion, grades the completion based on the prompt and criterion. 
        Below is a prompt, a completion, and a criterion with which to grade the completion. You need to respond according 
        to the criterion instructions. For reference, these are documents from the year \textcolor{blue}{\{year\}}. 
        \textcolor{blue}{\{strict\_phrase\}} \\
        
        \vspace{0.25cm}
        
        \textbf{Output:} \textcolor{blue}{\{narrative\_text\_of\_the\_company\}} \\
        
        
        \textbf{Criterion:} \textcolor{blue}{\{criterion\}} \\ \hline
    \end{tabular}
    \caption{The base prompt template, We replace the colored slot with real text before querying the LLMs.  (see table 4 and 5 for more details)}
    \label{tab:my_label}
\end{table}

\begin{table*}[h!]
    \centering
    \begin{tabular}{|c|p{12cm}|}
        \hline
        \textbf{Rating} & \textbf{Criterion for the Absolute Rating} \\ \hline

        \textbf{Confidence Rating} & \textbf{\#\# Criterion} \newline
        You should give the text a decimal numeric grade between 0 and 2. \newline
        2. The text is confident about robust growth and of greater returns next financial year. \newline
        1. Text in this category gives strong likelihood of company stability, but is either relatively unsure about future growth or not confident about it. \newline
        0. Text in this category shows that the company is not very robust, uncertain about its future, and most importantly, shows inconsistent and bad finances. \newline
        \textit{Answer only with a decimal number in the 0-2 range.} \\ \hline

        \textbf{Environmental Rating} & \textbf{\#\# Criterion} \newline
        You should give the text a decimal numeric grade between 0 and 2. \newline
        2. The text offers actionable plans relating to environment and sustainability and includes sustainability as a central goal. \newline
        1. Text in this category mentions commitments to sustainability and environment, but doesn't offer many actionable plans for the same. \newline
        0. Text in this category doesn't mention the environment and sustainability at all. \newline
        \textit{Answer only with a decimal number in the 0-2 range.} \\ \hline

        \textbf{Innovation Rating} & \textbf{\#\# Criterion} \newline
        You should give the text a decimal numeric grade between 0 and 2. \newline
        2. The text shows future plans as well as actions that the company has taken towards greater innovation in its operations. It also mentions a working R\&D unit. \newline
        1. Text mentions commitment to improve its practices and operations, and mentions innovation. However, there is little to no work done for the same in this current document. \newline
        0. Text emphasizes continuing its operations next year in the same manner, without any new innovations. \newline
        \textit{Answer only with a decimal number in the 0-2 range.} \\ \hline

        \textbf{People Rating} & \textbf{\#\# Criterion} \newline
        You should give the text a decimal numeric grade between 0 and 2. \newline
        2. The text acknowledges the importance of people and talent in driving the company forward. It also mentions actionable plans to attract new talent and ensure employee welfare. \newline
        1. Text mentions importance of people and talent to its operations but doesn't mention any employee welfare activities. \newline
        0. Text makes no mention of employee welfare and importance of people talent. \newline
        \textit{Answer only with a decimal number in the 0-2 range.} \\ \hline
    \end{tabular}
    \caption{The “Criterion” of evaluation of each rating type.}
    \label{tab:my_label}
\end{table*}

\begin{table*}[h]
    \centering
    \begin{tabular}{|c|c|}
        \hline
        \textbf{Name of Additional Text} & \textbf{Additional Text Added to Base Prompt} \\ \hline

        \textbf{narrative\_text\_of\_the\_company} & The actual text which is input from the filing. \\ \hline

        \textbf{strict\_phrase} & Miscellaneous changes to generate the “strict” prompt. \\ \hline

        \textbf{year} & Year of the SEC Filing. \\ \hline

    \end{tabular}
    \caption{Details of Additional Text Added to the Base Prompt}
    \label{tab:additional_text}
\end{table*}

\subsection{Relative Analysis.}
Similarly, we utilize pairwise comparison to compare narrative text of a particular section from multiple companies. The LLM is instructed to select one of the input excerpts as the best. The “win” is recorded. See table 6 for the prompt. In our analysis, we use three companies, “Royal Gold”, “IBM” and “Apple”. The results of this framework are presented in the next section.

\begin{table}[h]
    \centering
    \begin{tabular}{|p{8cm}|}
        \hline
        \textbf{Base Prompt for Relative Analysis} \\ \hline
        You are an AI grader that, given an output and a criterion, grades the completion based on the prompt and criterion. 
        Below "Excerpt A", "Excerpt B", and "Excerpt C", you must compare all excerpts and output which excerpt is better. \\
        
        \textbf{\#\# Excerpt A} \\
        \textcolor{blue}{\{excerpt\_1\}} \\

        \textbf{\#\# Excerpt B} \\
        \textcolor{blue}{\{excerpt\_2\}} \\

        \textbf{\#\# Excerpt C} \\
        \textcolor{blue}{\{excerpt\_3\}} \\

        \textbf{\#\# Criterion} \\
        Do not focus on the grammar; instead, focus on the overall future plan and robust explainability. \\

        [Answer with either "A", "B", or "C"] \newline
        A. If Excerpt A is the best, detailed, transparent with robust financials. \newline
        B. If Excerpt B is the best, detailed, transparent with robust financials. \newline
        C. If Excerpt C is the best, detailed, transparent with robust financials. \\
        \hline
    \end{tabular}
    \caption{Prompt template for relative analysis where “Excerpt” refers to the company’s text of a section}
    \label{tab:my_label}
\end{table}

\section{Results.}
To ensure robustness of the system, we tested it against three major publicly listed companies. Royal Gold (Ticker: \$RGLD), IBM (Ticker: \$IBM) and Apple (Ticker: \$AAPL).

\subsection{Absolute Analysis.} 
The result of the absolute analysis is a yearly rating between 0-2 of all companies. Based on the availability of documents, some ratings extend much further back in history than other ratings. For visualization, the results of the “strict” grader and the normal grader are averaged. The difference between their distributions for all companies is depicted in figure 2. Further, we ensure each criterion returns average results that are largely uncorrelated \cite{Willig2023} \cite{Elal2023} as shown in figure 3.

\begin{figure}[h]
    \centering
    \includegraphics[width=0.9\linewidth]{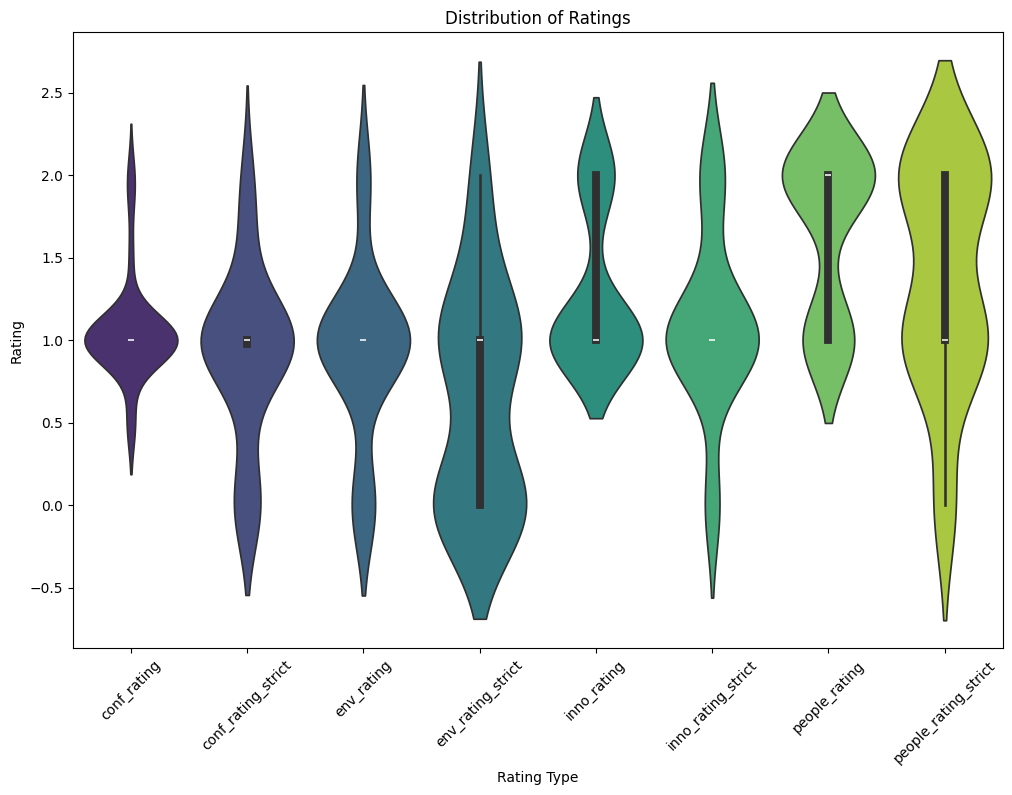}
    \caption{Range of the various categories of ratings}
    \label{fig:enter-label}
\end{figure}

\begin{figure}[h]
    \centering
    \includegraphics[width=0.9\linewidth]{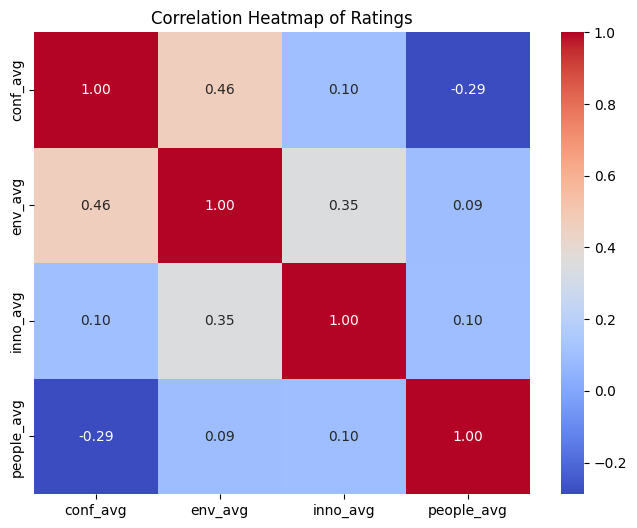}
    \caption{Correlation between the average ratings of all categories}
    \label{fig:enter-label}
\end{figure}

We represent the evolving priorities within each company over the years by visualizing the proportion of ratings in the categories of confidence, environment, innovation, and people, and how they change over the years. By analyzing the shifts in these ratings, we can infer how these companies have adjusted their strategies over time in response to both internal and external factors. 
\\

 \begin{figure}[h!]
     \centering
     \includegraphics[width=1\linewidth]{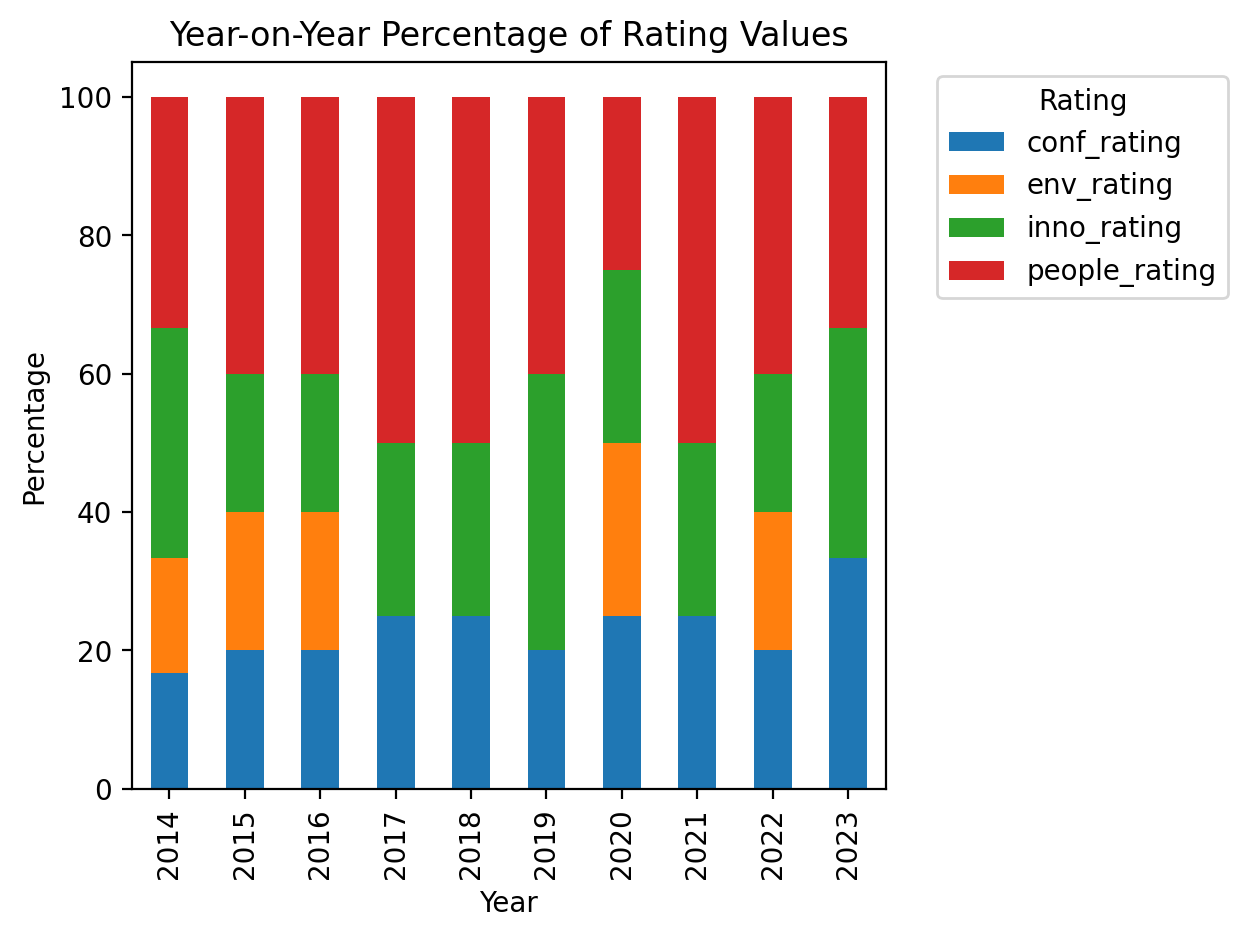}
     \caption{Representation of Year-on-Year priorities of Apple (\$AAPL)}
     \label{fig:enter-label}
 \end{figure}
\begin{figure}[h]
    \centering
    \includegraphics[width=1\linewidth]{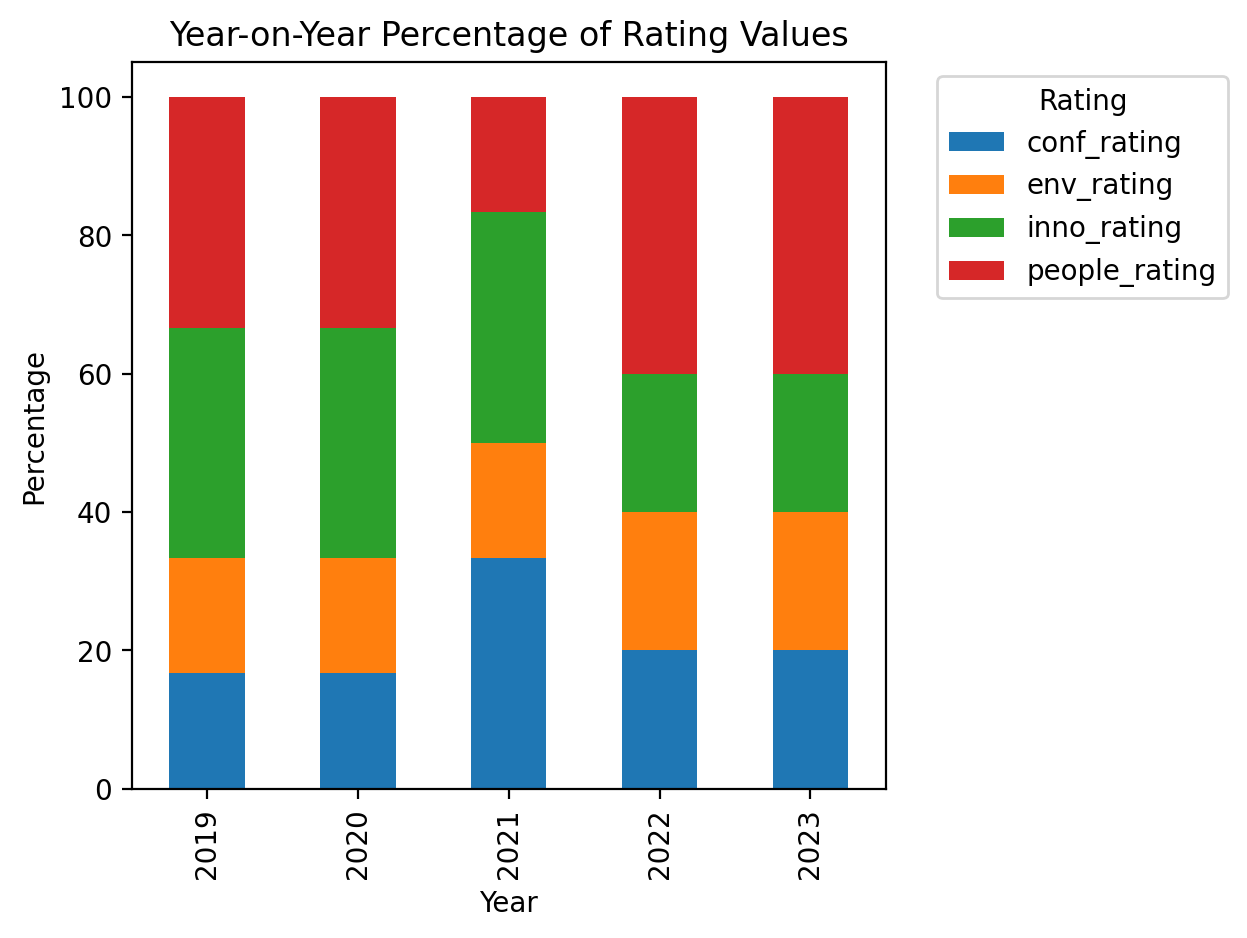}
    \caption{Representation of Year-on-Year priorities of IBM (\$IBM)}
    \label{fig:enter-label}
\end{figure}
\begin{figure}[h]
    \centering
    \includegraphics[width=1\linewidth]{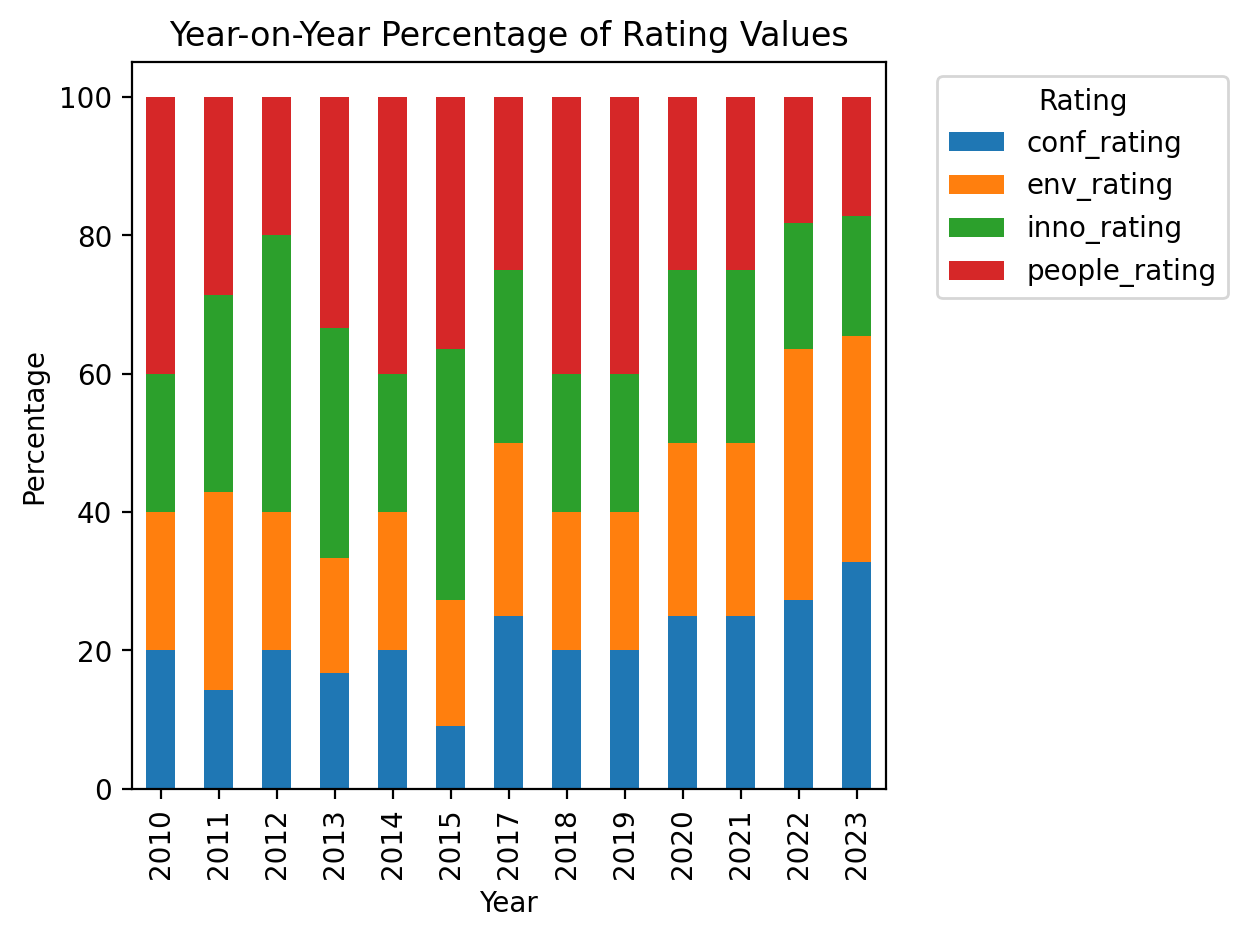}
    \caption{Representation of Year-on-Year priorities of Royal Gold (\$RGLD)}
    \label{fig:enter-label}
\end{figure}

\subsection{Relative Analysis.}

For six sections of the 10-K filing, we perform relative analysis with all three companies and record the LLM ratings. In this context, a "win" is defined as a scenario where a company outperforms its peers in a specific category for a given year. The categories are recognized as individual sections of the 10-K filing. We aggregate the number of wins across the sections, presenting a clear preference of which company has been the most successful in dominating the different categories over time in Figure 7. 

\begin{figure}[h!]
    \centering
    \includegraphics[width=1\linewidth]{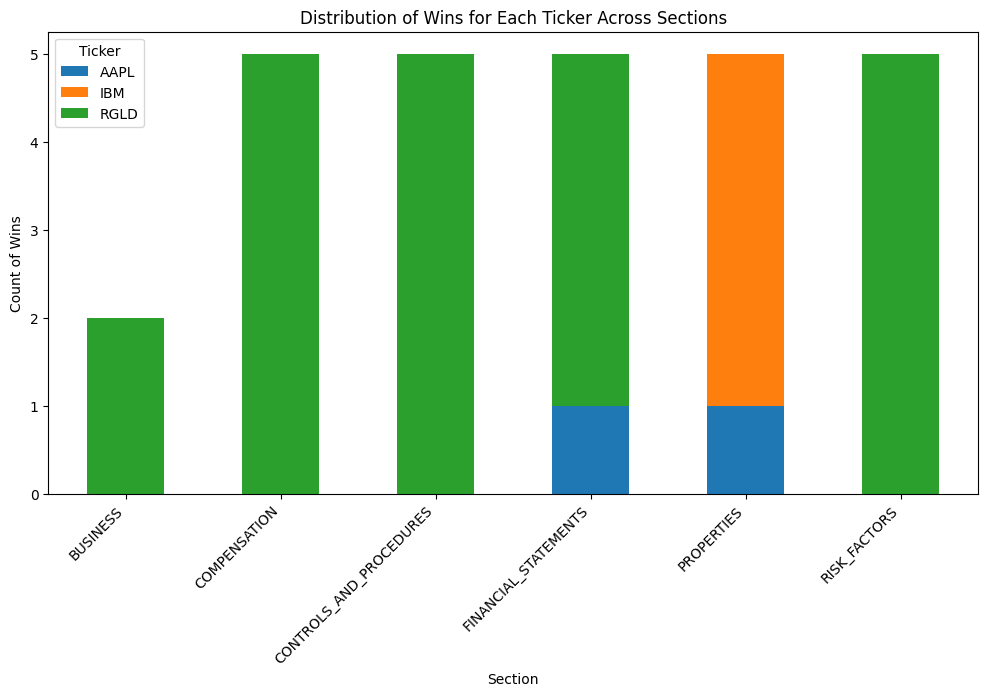}
    \caption{Relative analysis over multiple sections}
    \label{fig:enter-label}
\end{figure}

\subsection{Web Application.}
The demo web application is designed to facilitate the analysis and visualization of ratings derived from 10-K filings of publicly traded companies using large language models (LLMs). The application provides an interactive no-code interface where the end-user can input a company's stock ticker symbol and obtain a comprehensive analysis of the company's annual reports.

Upon entering a ticker symbol, the application retrieves and processes the relevant 10-K filings, either from a pre-cached dataset or through real-time data processing. The core functionality of the application revolves around generating and displaying the absolute ratings for the four key areas of interest. These ratings are derived from the text of the 10-K filings, and similar to the methodology followed throughout, the application employs two role-based agents—one with a standard assessment criterion and another with a stricter evaluation approach—to produce a range of rating for the absolute analysis.

The user interface of the application allows for interactive filtering of 10-K sections to be included in the analysis. The end-user can select which sections of the filings they wish to exclude, thus tailoring the analysis to specific areas of interest. Once the analysis is performed, the application generates the visual representations of the ratings. This streamlit based application is hosted on the lightning.ai cloud.

\section{Known Limitations.}
Despite the strengths of the 10-K Filings Rating System, several limitations were identified during the development process. One notable challenge is the difficulty in accurately identifying and categorizing all relevant sections of the 10-K filings. While the system successfully extracts a significant amount of valuable narrative text, some sections are not as meaningfully captured, leading to potential gaps in the analysis. Another critical limitation is the inherent bias that large language models (LLMs) can introduce. LLMs are trained on vast amounts of data, and as a result, they can inadvertently reflect biases present in that data, which might skew the ratings in unintended ways. This could impact the fairness and objectivity of the analysis, particularly in areas like environmental impact or employee relations. These limitations suggest that while the current implementation provides robust insights, there is room for further refinement to enhance the accuracy, fairness, and comprehensiveness of the analysis.

\section{Conclusion.}
In an era where the volume and complexity of financial data are ever-increasing, our research introduces a novel approach to the systematic analysis of SEC 10-K filings using large language models (LLMs). By automating the extraction, segmentation, and evaluation of these filings, our method offers a scalable solution that transforms qualitative corporate disclosures into actionable, quantitative ratings. This innovative approach not only streamlines the assessment of company performance across multiple dimensions—such as confidence, environmental sustainability, innovation, and workforce management—but also provides a cost-effective alternative to traditional, time-intensive financial analysis.

The integration of Cohere's Command-R+ LLM into our data pipeline enables the rapid processing and rating of large numbers of companies, offering stakeholders a powerful tool for making informed decisions with unprecedented efficiency. Moreover, our system’s focus on longitudinal analysis through year-on-year comparisons provides a more nuanced understanding of a company’s strategic evolution, allowing for a more accurate and fair assessment of its performance over time.

The development of a user-friendly, no-code web application further enhances the accessibility of our method, democratizing the ability to conduct sophisticated financial analyses without requiring deep technical expertise. As the financial landscape continues to evolve, our approach represents a significant advancement in the application of AI to corporate analysis, providing market analysts, investors, and other stakeholders with a robust, scalable tool for navigating the complexities of modern financial markets.
\\

\end{document}